\documentclass[journal]{IEEEtran}
\usepackage{amsmath,amsfonts}
\usepackage{algorithmic}
\usepackage{algorithm}
\usepackage{array}
\usepackage[caption=false,font=normalsize,labelfont=sf,textfont=sf]{subfig}
\usepackage{textcomp}
\usepackage[pagebackref=true,breaklinks=true,colorlinks,bookmarks=false]{hyperref}
\usepackage{stfloats}
\usepackage{url}
\usepackage{verbatim}
\usepackage{graphicx}
\usepackage{cite}
\usepackage{soul}
\usepackage{multirow}
\usepackage{amssymb}
\usepackage{xcolor}
\usepackage{booktabs}
\usepackage{makecell}
\usepackage{xspace}

\newcommand{\mkr}[1]{{\color{red}{{#1}}}}
\newcommand{\mkg}[1]{{\color{green}{{#1}}}}
\newcommand{\mkb}[1]{{\color{blue}{{#1}}}}
\begin{document}

\title{Co-attention Propagation Network for Zero-Shot Video Object Segmentation}

\author{Gensheng Pei,
        Yazhou~Yao,
	    Fumin~Shen,
	    Dan Huang,
	    Xingguo Huang,
	    and~Heng-Tao Shen (Fellow, IEEE)
	\thanks{G. Pei and Y. Yao are with the School of Computer Science and Engineering, Nanjing University of Science and Technology, Nanjing 210094, China.}
	\thanks{D. Huang is with the Chinese Academy of Ordnance Science, Beijing 100089, China.}
	\thanks{X. Huang is with the College of Instrumentation and Electrical Engineering, Jilin University, Changchun 130061, China.}
	\thanks{F. Shen and H. Shen are with the School of Computer Science and Engineering, University of Electronic Science and Technology of China, Chengdu 611731, China.}	
}

\markboth{}%
{Shell \MakeLowercase{\textit{et al.}}: A Sample Article Using IEEEtran.cls for IEEE Journals}


\maketitle

\begin{abstract}
Zero-shot video object segmentation (ZS-VOS) aims to segment foreground objects in a video sequence without prior knowledge of these objects.
However, existing ZS-VOS methods often struggle to distinguish between foreground and background or to keep track of the foreground in complex scenarios.
The common practice of introducing motion information, such as optical flow, can lead to overreliance on optical flow estimation.
To address these challenges, we propose an encoder-decoder-based hierarchical co-attention propagation network (HCPN) capable of tracking and segmenting objects.
Specifically, our model is built upon multiple collaborative evolutions of the parallel co-attention module (PCM) and the cross co-attention module (CCM).
PCM captures common foreground regions among adjacent appearance and motion features, while CCM further exploits and fuses cross-modal motion features returned by PCM.
Our method is progressively trained to achieve hierarchical spatio-temporal feature propagation across the entire video.
Experimental results demonstrate that our HCPN outperforms all previous methods on public benchmarks, showcasing its effectiveness for ZS-VOS.
Code and pre-trained model can be found at~\url{https://github.com/NUST-Machine-Intelligence-Laboratory/HCPN}.
\end{abstract}


\section{Introduction}\label{sec:1}
\IEEEPARstart{V}{ideo} object segmentation (VOS) is the task of distinguishing primary foreground objects from the background in each video frame. It has become a fundamental task in several fields, including video inpainting~\cite{lao2021flow}, background matting~\cite{sun2021deep}, and robotics~\cite{gowda2020alba}. Existing VOS methods have mainly focused on zero- and one-shot scenarios. The primary difference between them is whether or not the inference stage provides a pixel-level mask for the first frame. This paper focuses specifically on the challenging zero-shot VOS task, which automatically segments objects of interest without requiring manual interaction.

\begin{figure}[ht]
	\begin{center}
		\includegraphics[width=0.95\linewidth]{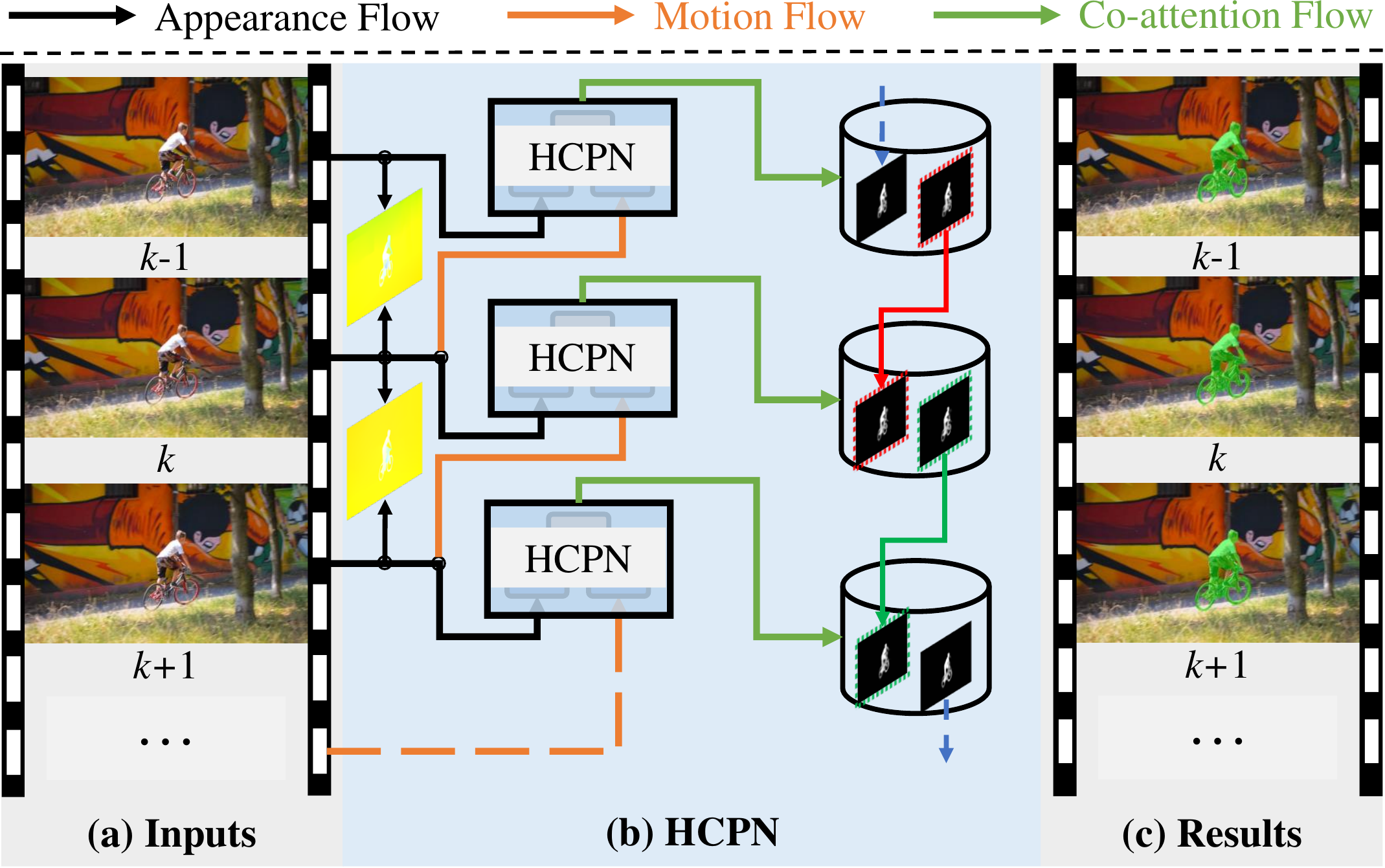}
	\end{center}
	\caption{\textbf{An illustration of HCPN}. \textbf{(a)} We compute the corresponding optical flow for adjacent frames. \textbf{(b)} HCPN captures short-term motion information with optical flow and combines it with contextual appearance features in a hierarchy. The motion and appearance features complement each other to form a propagation model for video sequences. We use consecutive frames and their optical flow, enabling all frames except the first and last to act as propagation paths (as shown by the colorful arrows). \textbf{(c)} This work uses optical flow to guarantee temporal and spatial consistency for zero-shot VOS.}\label{fig.1}
\end{figure}

One-shot VOS approaches utilize the annotation of the first frame to identify primary objects. On the other hand, the main challenge with zero-shot VOS is that the primary objects are unknown beforehand. This lack of prior knowledge may result in ambiguity with segmented objects, making it challenging to segment identical objects. The current consensus among researchers is that the foreground objects to be segmented in the video sequence are located in salient regions, mostly in motion. To address zero-shot VOS, many existing models~\cite{MATNet,FSNet,AMCNet} obtain information about both appearance and motion. Moreover, several approaches~\cite{FSEG,ARP,AGNN} use unsupervised optical flow estimation and recurrent neural networks (RNNs) to model spatio-temporal representations, in order to analyze the relationship between primary objects and temporal sequences.

Although the above methods have achieved promising performance in zero-shot VOS, they still have some limitations. Optical flow can only estimate the motion information between two consecutive frames and accumulates estimation errors over time. To address this limitation, motion-specific appearance (MSA) based methods effectively guide motion information to extract the appearance features of moving objects. MSA indicates the specific range (or region) of appearance features determined based on motion information.
On the other hand, recurrent neural network-based approaches have computationally demanding limitations in building long-term models. RNNs-based methods aim to explore the information about each frame's appearance features over time and establish the connection between appearance and motion, and are thus considered appearance-specific motion-based (ASM) methods. ASM matches specific moving objects by the similarity of their appearance features.
However, a single MSA- or ASM-based approach cannot effectively address this challenge when the background appearance is intricate and similar to the foreground appearance (Fig.~\ref{fig.1} \textbf{(a)}). The reasons for this are: \textbf{1)} MSA relies heavily on the guidance information of optical flow, and the estimation error could affect the segmentation performance. \textbf{2)} ASM draws on similar inter-frame features, so moving objects could cause considerable feature differences.
In this work, we aim to weaken the model's overreliance on optical flow and ensure that similarities between successive frames are more accurate.

\begin{figure}[t]
	\begin{center}
		\includegraphics[width=0.95\linewidth]{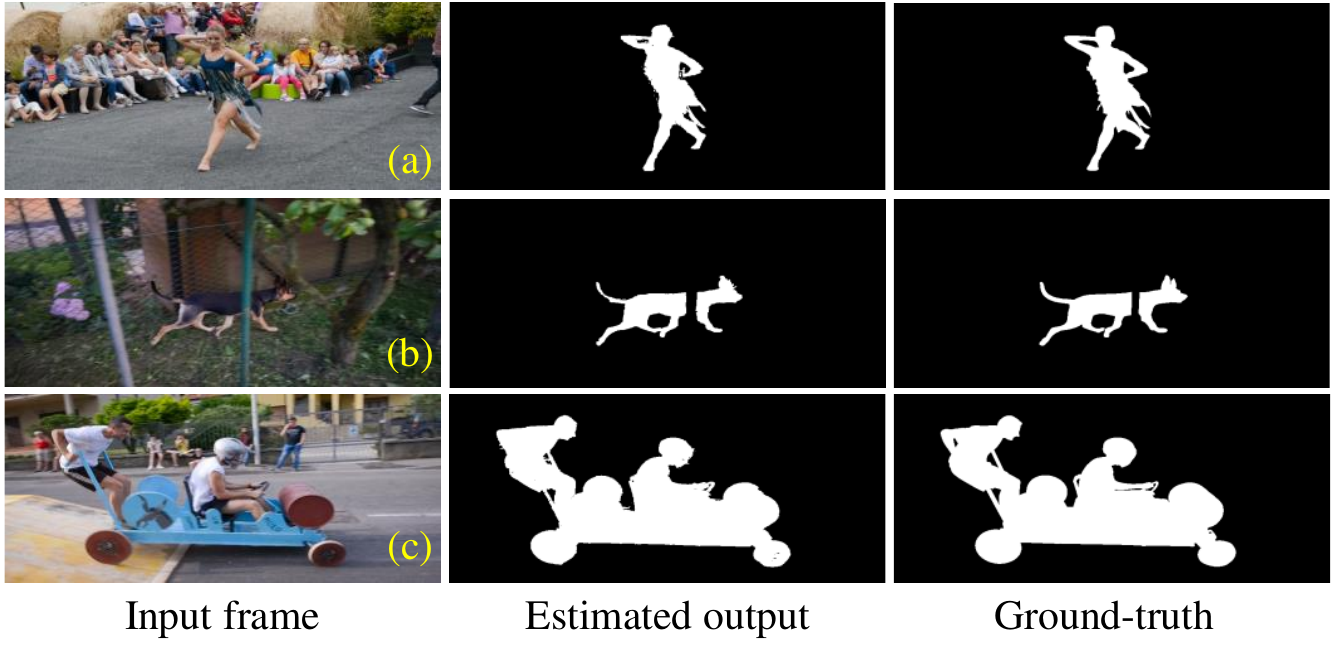}
	\end{center}
	\caption{Examples on (a) background clutter with a single foreground object, (b) motion object video with background occlusion, and (c) fast motion with multiple objects. Our method yields accurate estimated segmentation masks.}\label{fig:2}
\end{figure}

To facilitate and mutually guide appearance and motion information, it is necessary to design a learning scheme. This paper proposes a hierarchical co-attention propagation network (HCPN) that establishes a contextual connection between motion-specific appearance and appearance-specific motion (Fig.~\ref{fig.1} \textbf{(b)}). 
Firstly, motion information is extracted in the short term by estimating the optical flow of two consecutive frames. To enable information exchange between MSA and ASM features, a hierarchical co-attention mechanism is designed with two main modules: the parallel co-attentive module (PCM) and the cross co-attention module (CCM). 
The PCM employs optical flow as prior knowledge to acquire more effective appearance features, while the appearance features complement the local information about the foreground that is missing from the optical flow.
Meanwhile, the CCM fuses the multimodal features returned by the PCM between inter-frames. To alleviate the negative impact of noise in some video frames, a propagation mechanism for the target foreground is established in all intermediate frames except the first and last frames. HCPN, equipped with the above merits, achieves desirable segmentation performance (Fig.~\ref{fig.1} \textbf{(c)}).

Optical flow guided zero-shot VOS methods have demonstrated significantly superior segmentation performance. Previous methods~\cite{MATNet,AMCNet,FSNet,RTNet,DBSNet,pei2022hierarchical} focus on obtaining higher-quality optical flow estimates corresponding to video frames.
However, forgoing appearance modeling by converting VOS to foreground motion prediction based on optical flow information does not handle static foreground objects well.
Additionally, poor optical flow estimation directly affects the segmentation results of the final model when faced with occlusion, motion blur, fast-moving objects, or even stationary objects in zero-shot VOS.
In such cases, the non-selective fusion of appearance and motion features may compromise segmentation accuracy. Moreover, when dealing with more complex and realistic video sequences, the slow motion of the target object may not allow for reliable optical flow estimation.
Our proposed HCPN in this paper addresses the above-mentioned problems through hierarchical feature representation, selection, and fusion steps. Two consecutive frames and their corresponding optical flow ensure that segmentation performance is guaranteed using appearance features even when the motion information fails.

Figure~\ref{fig:2} shows that the proposed method achieves satisfactory segmentation results on some challenging examples. To validate the effectiveness of HCPN, we conducted experiments on three widely used VOS benchmark datasets: DAVIS~\cite{DAVIS-16,DAVIS-17}, YouTube-Objects~\cite{YouTube-Objects}, and FBMS~\cite{FBMS}. Extensive results demonstrate that HCPN achieves optimal performance on object-level zero-shot VOS tasks compared to existing state-of-the-art (SOTA) methods, and it ranked first in the DAVIS leaderboard for the instance-level task.

The main contributions of this paper are summarized as follows: \textbf{1)} We propose a new architecture called the hierarchical co-attention propagation network, which automatically explores inter- and intra-frame representations of motion and appearance features. This architecture benefits from the input structure of two frames and intermediate optical flow to reduce the model's dependence on optical flow. \textbf{2)} The hierarchical evolution of PCM and CCM in our network enables the optical flow to act as a bridge between frames, propagating the primary objects in the video. \textbf{3)} Based on the modules mentioned above, our method outperforms existing SOTAs in several zero-shot VOS datasets and provides a more straightforward way to implement the instance-level zero-shot VOS setting.

\section{Related Work}\label{sec:2}

\subsection{Video Object Segmentation}\label{sec:2.1}

VOS is roughly classified into Zero-Shot VOS (also known as unsupervised VOS) and One-Shot VOS (also known as semi-supervised VOS). These categories are defined based on the type of supervision provided during the testing phase. 

\textbf{One-Shot Video Object Segmentation (OS-VOS)} methods utilize the annotated object mask in the first frame or keyframe to infer remaining frames. These methods can be broadly classified as propagation-based, detection-based, hybrid-based, and matching-based.
Propagation-based approaches~\cite{duke2021sstvos,ventura2019rvos} leverage temporal information to evolve an object mask propagator that refines the mask propagated from the previous frame. Detection-based approaches~\cite{lin2021query} create an object detector from the object appearance of the first frame and then crop out the object for segmentation. Hybrid-based approaches~\cite{SwiftNet,chen2020state} integrate both propagation and detection schemes, offering the advantages of both strategies. Matching-based approaches~\cite{yang2020collaborative,GMem,zeng2019dmm} typically train a Siamese matching network to find the pixel that best matches the past frame(s) and the current frame (or query frame) and accomplish the corresponding label assignment. Some matching-based works~\cite{STM,GMem,LVO,robinson2020learning} use a memory mechanism to retain previously segmented frame information, which enables learning of long-term spatial-temporal information and facilitates the evolution of objects over time.

\textbf{Zero-Shot Video Object Segmentation (ZS-VOS)} makes certain restrictive assumptions about the application scenario, thereby rendering manual annotation of the first frame unnecessary. In the initial stage, the primary objective of ZS-VOS is to automatically separate the main objects from the background. Earlier conventional methods relied on specific heuristics that were associated with the foreground (\textit{i.e.}, object proposals~\cite{fragkiadaki2015learning}, motion boundaries~\cite{FST}, saliency information~\cite{wang2015saliency}). These methods also required hand-crafted features, including color, contours, and optical flow. Based on object appearance and motion cues, several methods have been proposed to focus on the long-term motion information of point trajectories~\cite{EPONet,FBMS}, background subtraction~\cite{lu2020learning} and over-segmentation~\cite{DFNet} for ZS-VOS.

Following the successful application of deep learning in computer vision, many methods started using deep neural networks for ZS-VOS. Fully Convolutional Networks (FCN) were initially developed for image segmentation and later adapted to segment object(s) in videos. However, purely relying on optical flow information for foreground motion prediction, and abandoning appearance modeling, does not handle static foreground objects effectively. To address this issue, several methods proposed the two-stream structure~\cite{FSEG,LVO} or a CNN-based encoder-decoder network~\cite{SFL} to fuse motion and appearance information for better performance.
IET~\cite{IET} learned foreground object localization in static images, while MATNet~\cite{MATNet} designed a motion-attentive transition network to determine the background by analyzing motion-to-appearance. To achieve temporal stability among frames, PDB~\cite{PDB} constructed a ConvLSTM architecture for video salient object detection, and 3DCSEG~\cite{3DCSEG} used the 3D CNN technique to address ZS-VOS with temporal stability. Additionally, AnDiff~\cite{AnDiff} used an aggregation technique to propagate the features of the anchor from the first frame to the current frame to enable image correlations.

To leverage the inherent correlations among continuous video frames, DFNet~\cite{DFNet} proposed a discriminative feature network that considers more information during object inference.
With the advancement of optical flow estimation in recent years, many optical flow-based ZS-VOS methods have achieved significant improvements. However, this paper introduces a new perspective by reconsidering the trade-off between motion and appearance features. 
Unlike previous optical flow-based approaches, we construct a three-stream network structure to attenuate the risk of optical flow failure.

The fundamental goal of both OS-VOS and ZS-VOS methods is to establish an effective approach for distinguishing between background and foreground. The most significant difference between video segmentation and image segmentation is that the objects to be segmented in a video sequence exhibit strong spatial and temporal consistency within that sequence. Hence, to develop a powerful segmentation model, we need to exploit the temporal and spatial information present in the video frames. Building upon these observations, in this study, we propose a three-stream encoder-decoder network for multimodal feature fusion and intra-frame object propagation. This approach aims to achieve MSA and ASM to summarize a more robust target representation.

\begin{figure}[t]
	\begin{center}
		\includegraphics[width=0.9\linewidth]{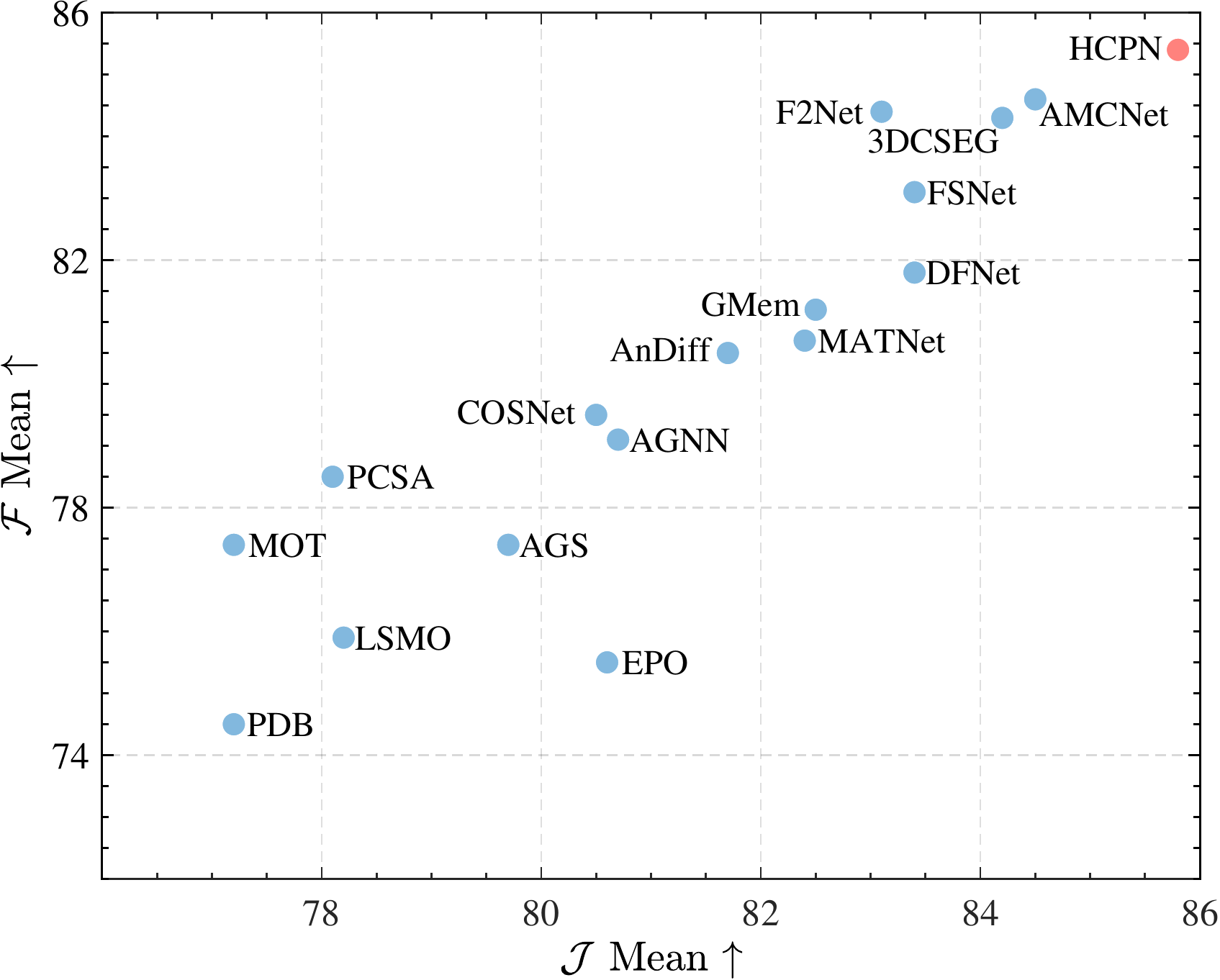}
	\end{center}
	\caption{Boundary accuracy Mean $\mathcal{F}$ versus region similarity Mean $\mathcal{J}$ on the DAVIS-16 validation set~\cite{DAVIS-16}. Existing and proposed methods are marked with {\large\textcolor[rgb]{0.51,0.72,0.87}{$\bullet$}} and {\large\textcolor[rgb]{1,0.51,0.49}{$\bullet$}}, respectively. Compared with the best reported ZS-VOS model, the proposed method HCPN achieves a new SOTA by a large margin.}\label{fig:3}
\end{figure}

\begin{figure*}[t]
	\begin{center}
		\includegraphics[width=1.0\linewidth]{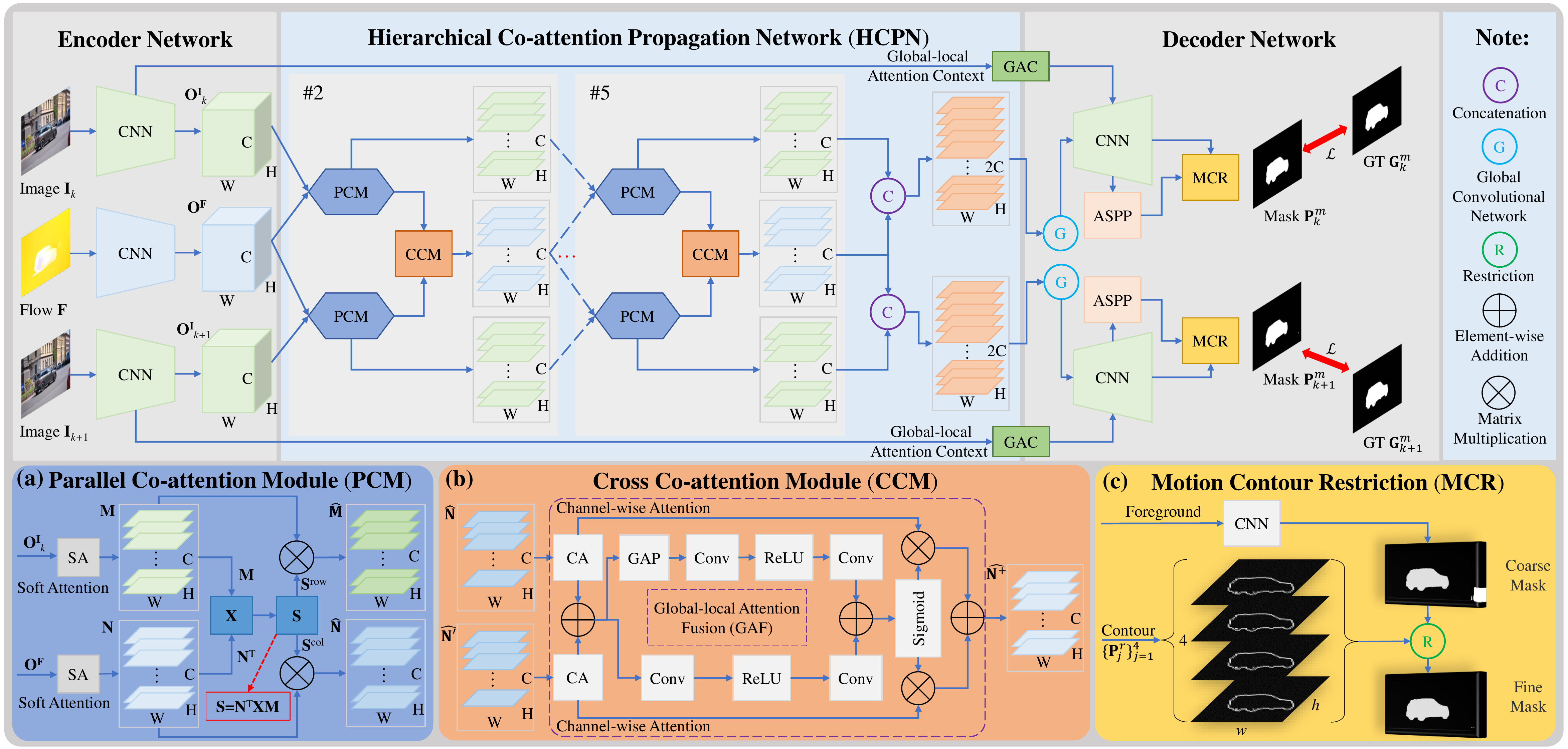}
	\end{center}
	\caption{The HCPN architecture consists of four cascaded HCPNs numbered \#2 to \#5. These take image frames ($\mathbf{I}_{k}$, $\mathbf{I}_{k+1}$) and optical flow ($\mathbf{F}$) as inputs to extract multimodal spatio-temporal feature maps. In each HCPN, \textbf{(a)} PCM is used to enhance the spatio-temporal representation of the feature maps, and \textbf{(b)} CCM focuses on efficiently fusing multimodal features. Finally, \textbf{(c)} the MCR module constrains the multi-scale contours $\{\mathbf{P}^r_j\}^4_{j=1}$ to coarse masks for refining masks. Please see the zoomed-in view for more details.}
	\label{fig:4}
\end{figure*}

\subsection{Attention Mechanism}\label{sec:2.2}

Attention mechanisms, which draw inspiration from human perception, have been widely utilized in recent neural network architectures and have demonstrated significant improvements in various tasks, \textit{e.g.}, instance segmentation~\cite{wang2020end,athar2020stem,qi2022occluded} and visual recognition~\cite{yang2020gated,srinivas2021bottleneck,Attention}. By selectively screening semantic feature information and focusing on the most salient parts based on visual attention characteristics, attention-based approaches have been able to achieve these improvements without increasing the number of parameters.

In the VOS task, attention-based approaches~\cite{CFAM,SwiftNet,lin2021query,duke2021sstvos,heo2021guided} are primarily used as a tool for specific feature extraction, whereas our motivation is fundamentally distinct from theirs. The attention mechanism is essentially a feature fusion approach, which aims to represent the intrinsic connections among video frames more effectively. Additionally, the feature maps of inter-frame images exhibit a hierarchical structure that ranges from coarse to fine in different backbone layers. Therefore, we leverage the attention mechanism's role in feature fusion through two attention modules: PCM and CCM, which enable the efficient fusion and propagation of multimodal features, \textit{i.e.}, appearance and motion, at each layer.

\section{Method}\label{sec:3}

In this section, we present an overview of the proposed method's architecture in \S{\ref{sec:3.1}}. We explain the hierarchical co-attention propagation mechanism in detail in \S{\ref{sec:3.2}}. Additionally, global-local attention context and motion contour restriction are introduced in \S{\ref{sec:3.3}} and \S{\ref{sec:3.4}}, respectively. Finally, the last section \S{\ref{sec:3.5}} shows detailed network architecture. Fig. \ref{fig:3} shows the comparison results of HCPN and 16 SOTA methods in terms of Mean $\mathcal{J}\&\mathcal{F}$.

\subsection{Network Structure}\label{sec:3.1}

In the ZS-VOS task, determining the target object in unseen videos without any prior information is crucial. Inferring motion foreground based on optical flow is an effective technique. Moreover, we are aware that the appearance carries more object feature information than optical flow. Specifically, optical flow provides hints for foreground determination, whereas the appearance features of video frames can better complement the local detail features that optical flow lacks. Based on these observations, we propose an end-to-end trainable network architecture called HCPN, illustrated in Fig.~\ref{fig:4}.

\textbf{Encoder Network.}
The encoder with two-stream or three-stream is effective in various tasks~\cite{yao2021jo,sun2022pnp,yao2017exploiting,sun2021webly,yao2016domain}. 
To fulfill our objective, we employ a three-stream network as a hierarchical encoder to extract spatial-temporal features from adjacent video frames.
The encoder takes two video frames $\mathbf{I}_i\in\mathbb{R}^{{w}\times{h}\times{3}}|_{i=k, k+1}$ and their corresponding optical flow $\mathbf{F}\in\mathbb{R}^{{w}\times{h}\times{3}}$ as input. First, the encoder produces three feature maps, which consist two appearance feature maps $\mathbf{O}^\mathbf{I}_i\in\mathbb{R}^{{W}\times{H}\times{C}}|_{i=k, k+1}$ and one motion feature map $\mathbf{O}^\mathbf{F}\in\mathbb{R}^{{W}\times{H}\times{C}}$, by utilizing three parallel convolutional network layers as the backbone network. Next, the multimodal feature maps are fused and enhanced by multiple cascaded HCPNs to obtain the final desired features. Denote $W$ and $H$ as the width, height of the feature maps, and $C$ is the number of feature channels. The detailed processing of the encoder is in \S\ref{sec:3.2}.

\textbf{Bridge Network.}
Feature maps with large receptive fields~\cite{gao2021global2local,yang2020gated} help avoid local ambiguities caused by feature maps with small receptive fields. This is because the global context attention can adjust the feature maps to fit objects of different sizes. Additionally, the local context attention effectively highlights the correct object regions and suppresses noise caused by redundant features. To efficiently select spatio-temporal information for the feature maps extracted by the encoder, we propose the global-local attention context (GAC) module. This module passes correlations among channels of the multimodal feature maps to the decoder. We describe the details of the GAC module in \S\ref{sec:3.3}.

\textbf{Decoder Network.}
To generate fine segmentation masks from coarse segmentation masks, the proposed method's decoder uses the Atrous Spatial Pyramid Pooling (ASPP)~\cite{chen2017deeplab} and Motion Contour Restriction (MCR) modules. The ASPP operates on the feature maps from the bridge network and expresses them at multiple scales to increase the object's perceptual field and provide more spatial information.
Skip-connections then exploit multi-scale features from different layers, which are merged to produce the final predicted segmentation mask and contour mask using a merging strategy.
Finally, the MCR module restricts the segmentation masks to the object's contour regions, producing the final fine masks. Please refer to \S\ref{sec:3.4} for the MCR's operation details.

\subsection{Hierarchical Co-attention Propagation}\label{sec:3.2}
Many current optical flow-based methods use one video frame and its corresponding optical flow as inputs to extract features that are then fused and passed to the decoder to produce the final segmentation result. However, this approach has a significant drawback: it is heavily reliant on the quality of optical flow estimation, which can easily lead to incorrect determinations of the video target.
To address this issue, it is essential to establish a reliable region for segmentation, even in the absence of an explicit target. To achieve this, we propose a three-stream encoder that takes two consecutive RGB images and their optical flow as inputs. The hierarchical encoding process can be expressed as
\begin{equation}\label{eq1}
	(\mathbf{V}_k,\mathbf{V}_{k+1}) = \mathcal{F}_\text{HCPN}(\mathbf{O}^\mathbf{I}_k, \mathbf{O}^\mathbf{F}, \mathbf{O}^\mathbf{I}_{k+1})\text{,}
\end{equation}
where the features after encoding by $\mathcal{F}_\text{HCPN}$ are denoted as $\mathbf{V}_{\cdot}\in\mathbb{R}^{{W}\times{H}\times{2C}}$.
$\mathcal{F}_\text{HCPN}$ consists of two main modules: the PCM and CCM modules, respectively.

\textbf{Parallel Co-attention Module.}
As depicted in Fig.~\ref{fig:4} (a), the PCM takes image and optical flow feature maps as input, which are then passed through multiple~\textit{Soft Attention} (SA) units~\cite{chen2016attention,jaderberg2015spatial} to highlight content-informative regions in the appearance-attentive ($\mathbf{M}\in\mathbb{R}^{{W}\times{H}\times{C}}$) or motion-attentive ($\mathbf{N}\in\mathbb{R}^{{W}\times{H}\times{C}}$) feature maps. 
To obtain multiple attention feature maps, we employ the feature maps' co-attention approach inspired by~\cite{nguyen2018improved}, attending to appearance and motion features in parallel. This results in the generation of appearance and motion co-attentive features denoted as ${\mathbf{\hat{M}}}\in\mathbb{R}^{{W}\times{H}\times{C}}$ and ${\mathbf{\hat{N}}}\in\mathbb{R}^{{W}\times{H}\times{C}}$, respectively.
\begin{equation}\label{eq2}
	\begin{split}
		\mathbf{S} = &\mathbf{N}^\mathrm{T}\mathbf{X}\mathbf{M} = \mathbf{N}^\mathrm{T}\mathcal{P}(\mathbf{X})\mathbf{M}\text{,} \\
		{\mathbf{\hat{M}}} = &\mathbf{M}\mathcal{S}_\text{row}(\mathbf{S})\text{, } {\mathbf{\hat{N}}} = \mathbf{N}\mathcal{S}_\text{col}(\mathbf{S})\text{,} 
	\end{split}
\end{equation}
where $\mathbf{X}\in\mathbb{R}^{{C}\times{C}}$ represents the learnable weight matrix, while $\mathbf{S}\in\mathbb{R}^{({W}{H})\times({W}{H})}$ denotes the affinity matrix between appearance and motion feature maps. The linear mapping of $\mathbf{X}$ to multiple low-dimensional spaces is represented by $\mathcal{P}$.
$\mathcal{S}_\text{row}$ and $\mathcal{S}_\text{col}$ indicate row-wise and column-wise \textit{softmax}, respectively.
By applying this approach, the feature of another frame can be obtained as 
${\mathbf{\hat{N}}'}\in\mathbb{R}^{{W}\times{H}\times{C}}$. It is worth noting that our proposed PCM follows a hierarchical approach similar to existing methods for extracting and fusing motion and appearance features.

\textbf{Cross Co-attention Module.}
The PCM module is utilized to obtain the motion feature maps of two consecutive frames, including additional appearance information. To incorporate multi-scale feature contexts between successive frames, we employ a cross co-attention module with a global-local attention fusion (GAF) operation, as shown in Fig.~\ref{fig:4} (b).

Our proposed approach, called CCM, differs from previous methods in that it takes two frames of fused features and generates a more reliable motion representation through global-local attention fusion (GAF). 
The process begins with channel-wise attention (CA)~\cite{chen2017sca}, which is used to acquire channel-level relationships for each feature map. Next, GAF is utilized to mine global and local contexts, resulting in the fused feature ${\mathbf{\hat{N}}^+}\in\mathbb{R}^{{W}\times{H}\times{C}}$, which is defined as
\begin{equation}\label{eq3}
	{\mathbf{\hat{N}}^+} = \mathcal{C}({\mathbf{\hat{N}}})\dotplus{\mathcal{C}({\mathbf{\hat{N}'}})}\text{,}
\end{equation}
where $\mathcal{C}$ is channel-wise attention operation, and $\dotplus$ indicates GAF operation, which can be seen as a special kind of feature merging.
Global attention is achieved by $Global Average Pooling\rightarrow Conv\rightarrow ReLU\rightarrow Conv$. And the local attention is realized by $Conv\rightarrow ReLU\rightarrow Conv$.
The fused features are enhanced with global and local features (see the purple dashed box in Fig.~\ref{fig:4} (b)), respectively.
Then, these fused features have added again, passed through the sigmoid function, and multiplied with the original matrix to obtain the final ${\mathbf{\hat{N}}^+}$.
Subsequently, ${\mathbf{\hat{N}}^+}$ is passed as spatio-temporal features to the next hierarchy of HCPN. Finally, the encoded features of two frames are formulated as $\mathbf{V}_k=concat\left({\mathbf{\hat{N}}^+},{\mathbf{\hat{N}}}\right)$ and $\mathbf{V}_{k+1}=concat\left({\mathbf{\hat{N}}^+},{\mathbf{\hat{N}}'}\right)$.

\begin{figure}[t]
	\begin{center}
		\includegraphics[width=1\linewidth]{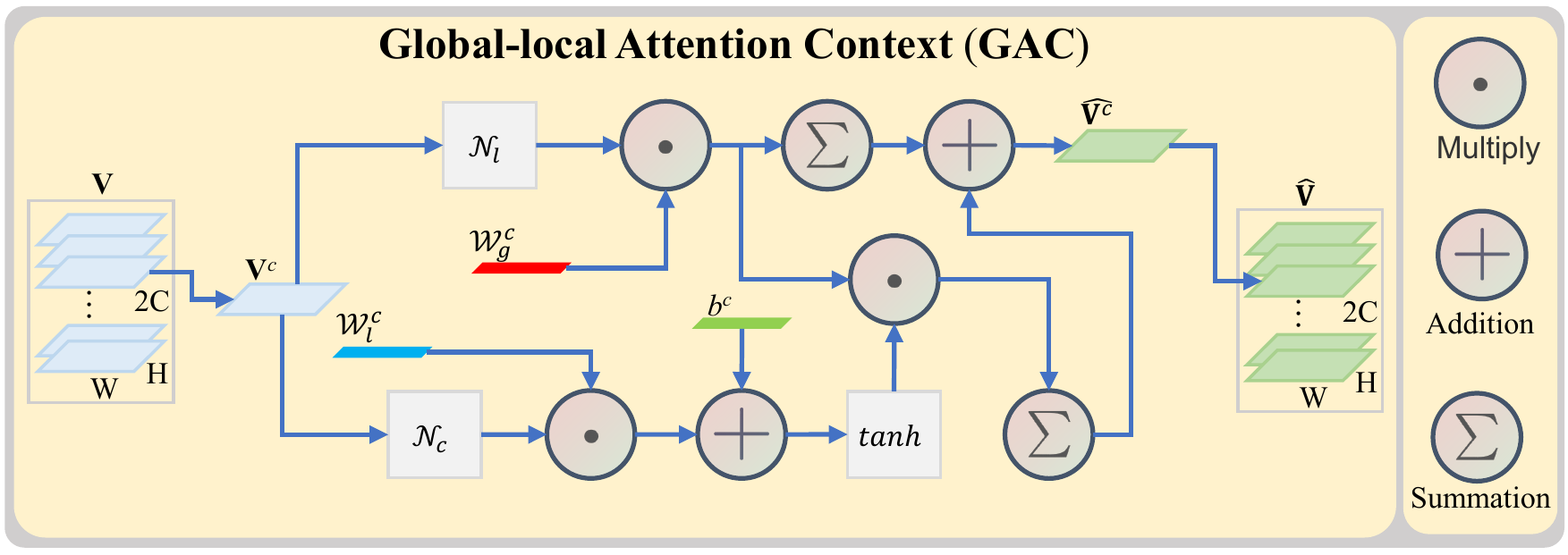}
	\end{center}
	\caption{Illustration of the proposed global-local attention context module. It allows splitting fused features into global and local contexts and merging them to enhance the fused feature representation.}\label{fig:5}
\end{figure}

\subsection{Global-local Attention Context}\label{sec:3.3}
The bridge network aims to embed motion features from the optical flow encoder stream and appearance features from the image encoder stream. To achieve this goal, we have developed the global-local attention context (GAC) module, as shown in Fig.~\ref{fig:5}, which integrates the feature maps from PCM and CCM. The global-attention context $\mathfrak{G}$ of GAC is focused on feature recalibration, while the local-attention context $\mathfrak{L}$ of GAC is used to enhance the correct object regions and remove redundant features or noise. 

For a spatio-temporal feature map $\mathbf{V}\in\mathbb{R}^{W\times{H}\times{2C}}$ of a frame, $\mathbf{V}$ represents the fused features obtained by PCM and CCM. The GAC module proposed in this work can split $\mathbf{V}$ into the global context (upper part of the figure) and local context (bottom part of the figure) separately and then merge them again to enhance the fused feature representation. This operation is similar to those utilized in semantic segmentation~\cite{chen2022saliency,yao2021non,10023953,chen2021semantically}. GAC is formulated as
\begin{equation}\label{eq4}
	\begin{split}
		{\mathbf{\hat{V}}} = \mathfrak{G}(\mathbf{V})&+{\mathfrak{L}(\mathbf{V})}\text{,} \\
		\mathfrak{G}(\mathbf{V}) = \sum_{c=1}^{2C} & \mathcal{W}^c_g\mathcal{N}_l\left(\mathbf{V}^c\right)\text{,} \\
		\mathfrak{L}(\mathbf{V}) = \sum_{c=1}^{2C} \mathcal{W}^c_g\mathcal{N}_l\left(\mathbf{V}^c\right)& tanh\big(\mathcal{W}^c_l\mathcal{N}_c\left(\mathbf{V}^c\right)+b^c\big)\text{.}
	\end{split}
\end{equation}
Here $\boldsymbol{\mathcal{W}}_g,\boldsymbol{\mathcal{W}}_l,\boldsymbol{b}\in\mathbb{R}^{C}$ are trainable parameters and $(\cdot)^c| c\in{\{1,2,\cdots,C\}}$ indicates each channel of $(\cdot)$. $\mathcal{N}_l$ and $\mathcal{N}_c$ represent $\ell_2$ normalization and channel normalization, respectively.

\subsection{Motion Contour Restriction}\label{sec:3.4}
In the decoder network, for each video frame, the model produces the coarse prediction mask $\mathbf{T}^m\in[0,1]^{w\times{h}}$ and its corresponding four contour masks $\{\mathbf{P}^r_j\in[0,1]^{w\times{h}}\}_{j=1}^4$ using ASPP~\cite{chen2017deeplab}.
These are intermediate results generated by the decoder. To obtain the background region, the contour masks are constrained using MCR. 
The final refined mask $\mathbf{P}^m\in[0,1]^{w\times{h}}$ is then produced using
\begin{equation}\label{eq5}
	\mathbf{P}^m = \mathcal{R}\left( \sum_{j=1}^4 \mathbf{P}^r_j \right) \circ \mathbf{T}^m \text{,}
\end{equation}
where $\circ$ represents the Hadamard product, and $\mathcal{R}$ can be considered a more dependable background reference obtained by combining the contour masks through voting.
The detailed pipeline of the proposed MCR module is illustrated in Fig.~\ref{fig:4} (c).
Finally, we employ the cross-entropy (CE) loss function as the optimization objective, which is represented as follows
\begin{equation}\label{eq6}
	\mathcal{L}=\mathcal{L}_{\rm{CE}}(\mathbf{G}^m,\mathbf{P}^m)+\mathcal{L}_{\rm{CE}}(\mathbf{G}^r,\mathbf{P}^r).
\end{equation}
Here, $\mathbf{G}^r$ denotes contour ground-truth and can be obtained through a binary mask.
The additional restriction $\mathcal{R}$ in Eq.~\eqref{eq5} utilizes $\mathcal{L}_{\rm{CE}}(\mathbf{G}^r,\mathbf{P}^r)$ to minimize mis-segmentation.

\subsection{Detailed Network Architecture}\label{sec:3.5}

\begin{figure*}[t]
	\begin{center}
		\includegraphics[width=1\linewidth]{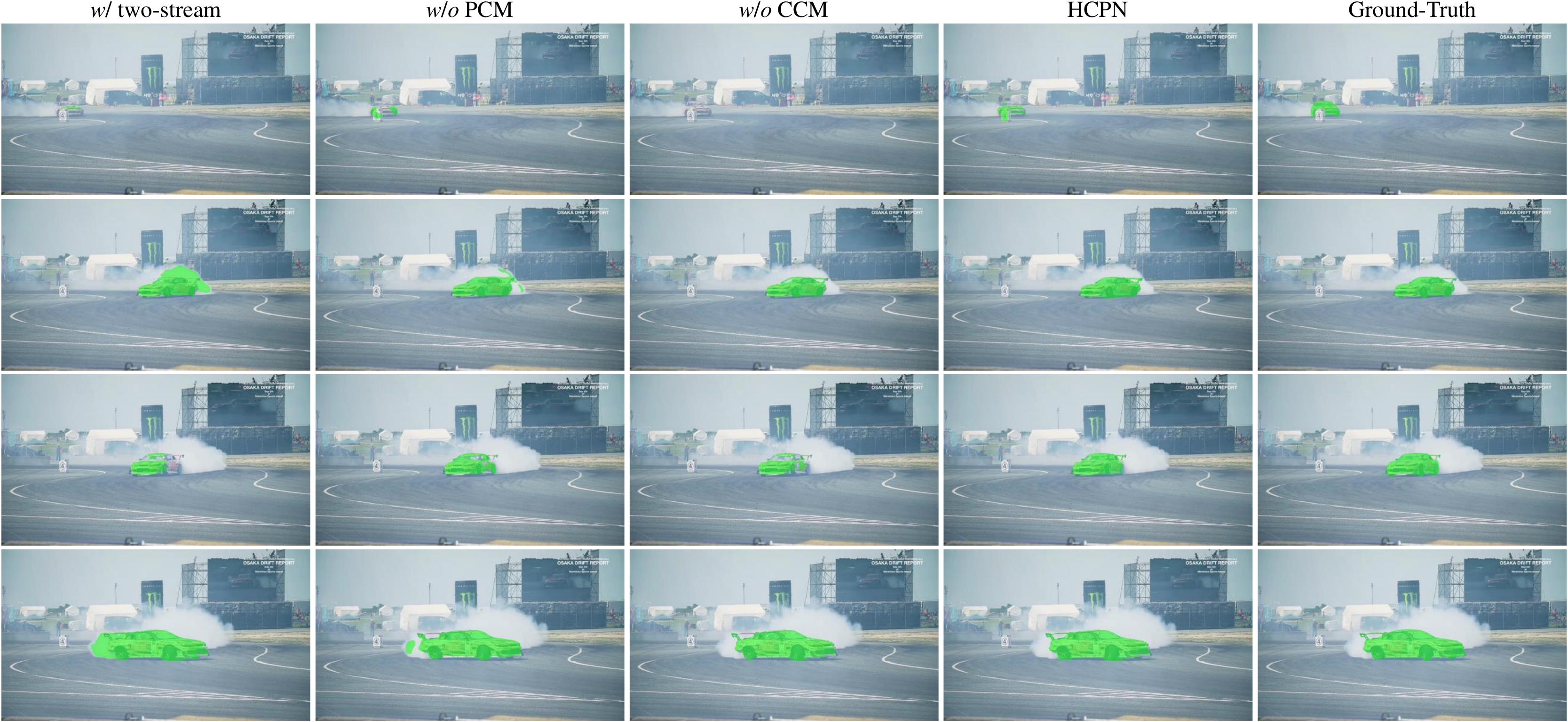}
	\end{center}
	\caption{Ablation study on sequence \textit{drift-chicane} from DAVIS-16. From left to right:  \textit{w}{/} two-stream,  \textit{w}{/}\textit{o} PCM,  \textit{w}{/}\textit{o} CCM, full HCPN and Ground-Truth.}\label{fig:6}
\end{figure*}

\textbf{Training Phase.}
Given two consecutive RGB frames $\mathbf{I}_i\in\mathbb{R}^{{512}\times{512}\times{3}}|_{i=k, k+1}$ in a video, RAFT \cite{RAFT} is employed to estimate the optical flow $\mathbf{F}\in\mathbb{R}^{{512}\times{512}\times{3}}$.
The ResNet-101~\cite{he2016deep} serves as the backbone network, which is fine-tuned using previous models~\cite{CFAM,DBSNet,AMCNet,MATNet_TIP}. We use PyTorch to implement the proposed method and conduct all experiments on a workstation with an Intel Xeon Gold 5118 CPU (2.30GHz) and two NVIDIA Tesla V100 32G GPUs.
We optimize the model parameters using the binary cross-entropy loss and stochastic gradient descent optimizer with a weight decay of 1e-5.
Specifically, the learning rates of the encoder and bridge networks are 1e-4, and that of the decoder network is 1e-3. The batch size is set to 10, and the entire training takes 62 hours for 25 epochs.

\textbf{Testing Phase.}
During the inference stage, we adjust the resolution of the unseen video frames and their corresponding optical stream to 512$\times$512, in accordance with the existing methods~\cite{AnDiff,DFNet,3DCSEG}.
The frames are then input into the trained network for video object segmentation. To obtain the final binary segmentation masks, we use CRF post-processing, following the standard strategy~\cite{AnDiff,AGS,AGNN}.
For each video frame, the proposed method takes 0.5 seconds for inference. The speeds of optical flow estimation and post-processing are 0.1 seconds per frame and 0.5 seconds per frame, respectively.
For more information on data properties and model implementation details, please refer to \S\ref{sec:4.1}.

\begin{figure*}[ht]
	\centering
	\includegraphics[width=0.49\linewidth]{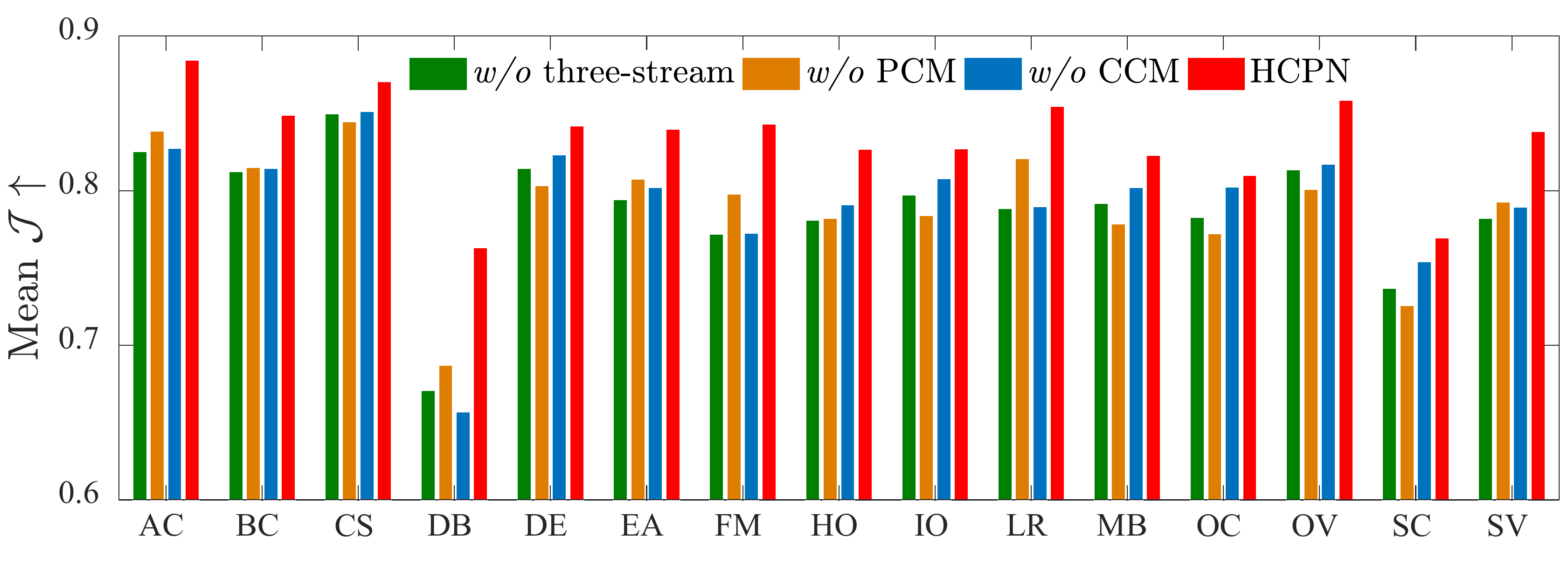}
	\label{4a}
	\centering
	\includegraphics[width=0.49\linewidth]{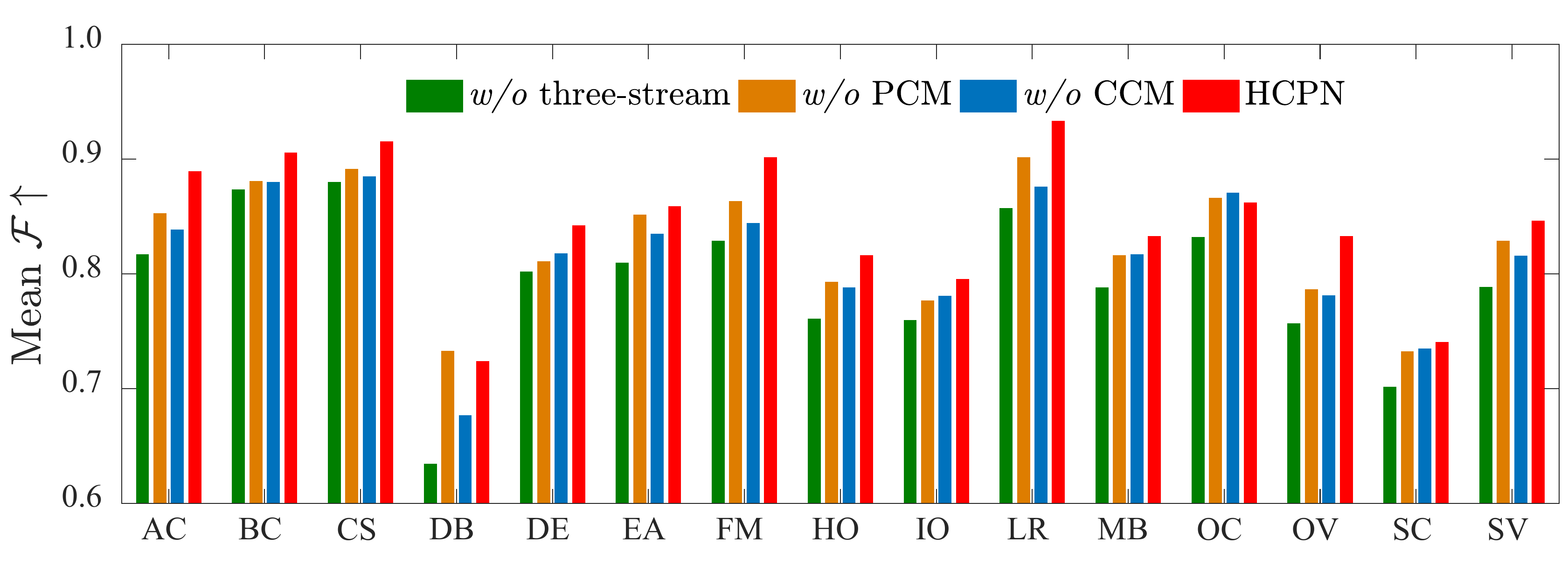}
	\label{4b}
	\centering
	\caption{Attitude-based study of the proposed various components on DAVIS-16. For each variant, we evaluate one metric (either Mean $\mathcal{J}$ or Mean $\mathcal{F}$) across sequences with specific attributes. The attributes include Appearance Change (AC), Background Clutter (BC), Camera Shake (CS), Dynamic Background (DB), Deformation (DE), Edge Ambiguity (EA), Fast Motion (FM), Heterogeneous Object (HO), Interacting Objects (IO), Low Resolution (LR), Motion Blur (MB), Occlusion (OC), Out-of-view (OV), Shape Complexity (SC), and Scale Variation (SV). These attributes are the same as those in Fig.~\ref{fig:10}.}\label{fig:7}
\end{figure*}

\section{Experiments}\label{sec:4}

\subsection{Experimental Setup}\label{sec:4.1}

\subsubsection{Datasets}\label{sec:4.1.1}

We conduct comprehensive experiments on three well-known datasets: the DAVIS~\cite{DAVIS-16,DAVIS-17}, YouTube-Objects~\cite{YouTube-Objects}, and FBMS~\cite{FBMS} benchmarks.

\textbf{DAVIS} is considered to be one of the most challenging datasets for video object segmentation, consisting of two subsets: DAVIS-16~\cite{DAVIS-16} and DAVIS-17~\cite{DAVIS-17}. DAVIS-16 comprises a total of 50 videos, with 30 for training and 20 for validation, and provides pixel-level annotations of foreground objects for each frame. In contrast, DAVIS-17 extends DAVIS-16 by including 70 additional videos, resulting in a total of 120 video sequences. The videos are divided into three sets: a training set of 60 videos, a validation set of 30 videos, and a test-dev set of 30 videos. To ensure a fair comparison on DAVIS-16, we employ the three standard protocols suggested by~\cite{DAVIS-16}: region similarity ($\mathcal{J}$), boundary accuracy ($\mathcal{F}$), and time stability ($\mathcal{T}$).

\textbf{YouTube-Objects}~\cite{YouTube-Objects} contains 126 web video sequences with 10 object categories, totaling more than 20,000 frames. 
To measure the segmentation performance on this dataset without further training, the region similarity $\mathcal{J}$ is used.

\textbf{FBMS} \cite{FBMS} consists of 59 videos, with 29 for training and 30 for testing.
Unlike DAVIS-like datasets, FBMS has sparsely labeled ground-truth annotations, with only 720 frames annotated across the entire dataset.
To follow the standard-setting, we do not fine-tune on the training set, and instead only evaluate segmentation performance on the test set using the region similarity metric $\mathcal{J}$.

\subsubsection{Implementation Details}\label{sec:4.1.2}

\begin{table}[t]
	\caption{Ablation study of input data. VF and OF indicate video frames and optical flow, respectively. $\oplus$, $\copyright$, and $\dotplus$ represent element-wise addition, concatenation, and GAF operations. `D' and `YT' denote DAVIS and YouTube-VOS.}\label{Tab:1}
	\begin{center}
		\begin{tabular}{|c||c||cc||cc|}
			\hline
			Component &Variant &Mean $\mathcal{J}\uparrow$ &$\Delta \mathcal{J}$ &Mean $\mathcal{F}\uparrow$ &$\Delta \mathcal{F}$\\
			\hline\hline
			\multirow{2}*{Input} &\textit{w}{/}\textit{o} OF &80.6 &-5.2 &80.9 &-4.5\\
			~ &\textit{w}{/}\textit{o} VF &80.8 &-5.0 &81.1 &-4.3\\
			\hline\hline
			\multirow{2}*{Stream} &\textit{w}{/} two &82.1 &-3.7 &81.2 &-4.2\\
			~ &\textit{w}{/} three &85.8 &- &85.4 &-\\
			\hline\hline
			\multirow{3}*{Fusion} &\textit{w}{/} {$\oplus$} &83.2 &-2.6 &82.9 &-2.5\\
			~ 					  &\textit{w}{/} {$\copyright$} &83.6 &-2.2 &83.1 &-2.3 \\
			~ 			 		  &\textit{w}{/} {$\dotplus$} &85.8 &- &85.4 &-\\
			\hline\hline
			\multirow{2}*{Optical Flow} &PWC &84.4 &-1.4 &83.9 &-1.5\\
			~ &RAFT &85.8 &- &85.4 &-\\
			\hline\hline
			\multirow{3}*{Dataset} &D &85.0 &-0.5 &84.8 &-0.6 \\
			                     ~ &YT &84.0 &-1.5 &83.7 &-1.7 \\
			                     ~ &D + YT &85.5 &- &85.4 &- \\
			\hline
		\end{tabular}
	\end{center}
\end{table}

The training procedure of the proposed method consists of three parts: 

\textbf{1)} Previous works often pre-train models on image datasets such as the salient object segmentation dataset MSRA10K~\cite{cheng2014global} and the instance segmentation dataset COCO~\cite{COCO}.
To extract more discriminative foreground features, the proposed method follows previous works~\cite{DBSNet,PMN,3DCSEG,MATNet_TIP} and uses the YouTube-VOS and DAVIS datasets as training samples. These datasets comprise 3,471+30 videos and a total of 14K frames.

\textbf{2)}
To fine-tune the proposed method for object-level segmentation setting, we use the 30 videos in the training set of DAVIS-16. Subsequently, we utilize the fine-tuned object-level model to validate the segmentation performance on DAVIS-16. The broad practicality of HCPN has been verified by its ability to perform well on the testing sets of both YouTube-Objects and FBMS without the need for further retraining.

\textbf{3)} 
To perform instance-level VOS, we first pre-trained the proposed model on the YouTube-VOS dataset and then optimized it on 60 video sequences from the training set of DAVIS-17. To adapt our object-level HCPN model for the instance-level VOS task, we introduced a foreground mask selection based on the Space-Time Memory (STM)~\cite{STM} method. In this paper, we followed a similar approach as~\cite{MSP} and used Mask R-CNN~\cite{MaskRCNN} to generate instance-level object masks for the first frame of each validation video. Next, we ran our model on the validation set of 30 videos from DAVIS-17 and generated object-level masks for each frame. Then, we used the primary object masks to remove background proposals and select the proposals from the foreground. Through these strategies, we transformed ZS-VOS into an OS-VOS setting and used STM for mask prediction.
Finally, we adopted the merging strategy proposed in~\cite{UnOVOST} to connect the instance-level segmentation masks performed by STM with the object-level segmentation masks obtained by our HCPN model, resulting in the final segmentation results.

\begin{table}[t]
	\caption{Ablation study of the proposed modules. Full indicates the full version of the proposed method.}\label{Tab:2}
	\vspace{-7pt}
	\begin{center}
		\begin{tabular}{|c||c||cc||cc|}
			\hline
			Component &Variant &Mean $\mathcal{J}\uparrow$ &$\Delta \mathcal{J}$ &Mean $\mathcal{F}\uparrow$ &$\Delta \mathcal{F}$\\
			\hline\hline
			\multirow{4}*{Module} &\textit{w}{/}\textit{o} PCM &81.9 &-3.9 &83.5 &-1.9\\
			~ &\textit{w}{/}\textit{o} CCM &82.7 &-3.1 &83.0 &-2.4\\
			~ &\textit{w}{/}\textit{o} GAC &84.7 &-1.1 &84.5 &-0.9\\
			~ &\textit{w}{/}\textit{o} MCR &85.2 &-0.6 &85.2 &-0.2\\
			\hline\hline
			Reference &Full &85.8 &- &85.4 &-\\
			\hline
		\end{tabular}
	\end{center}
	\vspace{-10pt}
\end{table}

\begin{table}[t]
	\caption{Ablation study of the efficacy on performance under different numbers of cascade HCPNs.}\label{Tab:3}
	\vspace{-7pt}
	\begin{center}
		\begin{tabular}{|c||c||cc||cc|}
			\hline
			Component &Variant &Mean $\mathcal{J}\uparrow$ &$\Delta \mathcal{J}$ &Mean $\mathcal{F}\uparrow$ &$\Delta \mathcal{F}$\\
			\hline\hline
			\multirow{3}*{HCPNs \#} &1 &81.1 &-4.7 &79.3 &-6.1 \\
			~ &2 &82.2 &-3.6 &81.4 &-4.0 \\
			~ &3 &84.2 &-1.6 &84.3 &-1.1 \\
			\hline\hline
			Reference &4 &85.8 &- &85.4 &-\\
			\hline
		\end{tabular}
	\end{center}
\end{table}

\subsection{Ablation Study}\label{sec:4.2}
To examine the effectiveness of the individual components and setups of HCPN, comprehensive ablation studies are performed on the DAVIS-16 validation set.

\subsubsection{Efficacy of feature Representation}
Our network uses two consecutive RGB frames and their corresponding optical flow as inputs.
The point at which the optical stream and RGB exchange information is precisely in the proposed PCM.
The results of the ablation study presented in Table~\ref{Tab:1} demonstrate that a single modal input (\textit{w}{/}\textit{o} OF or \textit{w}{/}\textit{o} VF) does not yield satisfactory results.
Table~\ref{Tab:2} shows that \textit{w}{/}\textit{o} PCM, which provides information interaction, results in a 3.9\% and 1.9\% decrease in the model's performance on Mean $\mathcal{J}$ and Mean $\mathcal{F}$, respectively.
Moreover, we examine the impact of varying the number of cascaded HCPNs on performance. The quantitative results in Table~\ref{Tab:3} indicate that the highest scores are achieved when HCPN is employed on all four convolution blocks.

\begin{table*}[t]
	\caption{The quantitative evaluation on the validation set of DAVIS-16 (see \S{\ref{sec:4.3.1}} for details). The results for other methods are borrowed from the public leaderboard. The three best scores are marked in \mkr{red}, \mkb{blue} and \mkg{green}, the same as in Tables \ref{Tab:5}, \ref{Tab:6}, \ref{Tab:7}.}\label{Tab:4}
	\begin{center}
		\begin{tabular}{|r||c||cc||ccc||ccc||c||c|}
			\hline
			\multirow{2}{*}{Methods~~} & \multirow{2}{*}{Publications} &  \multicolumn{2}{c||}{Operations} & \multicolumn{3}{c||}{$\mathcal{J}$} & \multicolumn{3}{c||}{$\mathcal{F}$} & $\mathcal{T}$    & $\mathcal{J}\&\mathcal{F}$ \\
			&  & OF & PP & Mean $\uparrow$ & Recall $\uparrow$ & Decay $\downarrow$ & Mean $\uparrow$ & Recall $\uparrow$ & Decay $\downarrow$ & Mean $\downarrow$ & Mean $\uparrow$ \\
			\hline\hline
			PDB~\cite{PDB} & ECCV-2018 &  & \checkmark & 77.2 & 90.1 & \mkb{0.9} & 74.5 & 84.4 & \mkr{-0.2} & 29.1 & 75.9 \\
			LSMO~\cite{LSMO} & IJCV-2019 & \checkmark & \checkmark & 78.2 & 89.1 & 4.1 & 75.9 & 84.7 & 3.5 & 21.2 & 77.1 \\
			MOT~\cite{MOT} & ICRA-2019 & \checkmark & \checkmark & 77.2 & 87.8 & 5.0 & 77.4 & 84.4 & 3.3 & 27.9 & 77.3 \\
			AGS~\cite{AGS} & CVPR-2019 & & \checkmark & 79.7 & 91.1 & \mkg{1.9} & 77.4 & 85.8 & 1.6 & 26.7 & 78.6 \\
			AGNN~\cite{AGNN} & ICCV-2019 & & \checkmark & 80.7 & 94.0 & \mkr{0.0} & 79.1 & 90.5 & \mkb{0.0} & 33.7 & 79.9 \\
			COSNet~\cite{COSNet} & CVPR-2019 & & \checkmark & 80.5 & 93.1 & 4.4 & 79.5 & 89.5 & 5.0 & \mkg{18.4} & 80.0 \\
			AnDiff~\cite{AnDiff} & ICCV-2019 & & &  81.7 & 90.9 & 2.2 & 80.5 & 85.1 & \mkg{0.6} & 21.4 & 81.1 \\
			EPO~\cite{EPONet} & WACV-2020 & \checkmark & \checkmark & 80.6 & 95.2 & 2.2 & 75.5 & 87.9 & 2.4 & 19.3 & 78.1 \\
			PCSA~\cite{PCSA} & AAAI-2020 & & &  78.1 & 90.0 & 4.4 & 78.5 & 88.1 & 4.1 & - & 78.3 \\
			MATNet~\cite{MATNet} & AAAI-2020 & \checkmark & \checkmark & 82.4 & 94.5 & 3.8 & 80.7 & 90.2 & 4.5 & 21.6 & 81.5 \\
			DFNet~\cite{DFNet} & ECCV-2020 & & \checkmark & 83.4 & 94.4 & 4.2 & 81.8 & 89.0 & 3.7 & \mkr{15.2} & 82.6 \\
			GMem~\cite{GMem} & ECCV-2020 & & \checkmark & 82.5 & 94.3 & 4.2 & 81.2 & 90.3 & 5.6 & - & 81.9 \\
			3DCSEG~\cite{3DCSEG} & BMVC-2020 & & & 84.2 & \mkg{95.8} & 7.4 & 84.3 & \mkg{92.4} & 5.5 & \mkb{16.8} & 84.2 \\
			F2Net~\cite{F2Net} & AAAI-2021 & & & 83.1 & 95.7 & \mkr{0.0} & 84.4 & 92.3 & 0.8 & 20.9 & 83.7 \\
			FSNet~\cite{FSNet} & ICCV-2021 & \checkmark & \checkmark & 83.4 & 94.5 & 3.2 & 83.1 & 90.2 & 2.6 & 21.3 & 83.3 \\
			AMCNet~\cite{AMCNet} & ICCV-2021 & \checkmark & \checkmark & \mkg{84.5} & \mkr{96.4} & 2.8 & \mkg{84.6} & \mkr{93.8} & 2.5 & - & \mkg{84.6}  \\
			CFAM~\cite{CFAM} & WACV-2022 & \checkmark &  & 83.5 & - & - & 82.0 & - & - & - & 82.8 \\
			DBSNet~\cite{DBSNet} & ACMMM-2022 & \checkmark & \checkmark & \mkr{85.9} & - & - & \mkb{84.7} & - & - & - & \mkb{85.3} \\
			\hline
			Ours & - & \checkmark & \checkmark & \mkb{85.8} & \mkb{96.2} & 3.4 & \mkr{85.4} & \mkb{93.2} & 3.0 & 18.6 & \mkr{85.6} \\ 
			\hline
		\end{tabular}
	\end{center}
\end{table*}

\subsubsection{Impact of Training Datasets}
To train our model, we adopt the same training data criteria as previous methods such as~\cite{DBSNet,AMCNet,3DCSEG,MATNet}, and use a consistent dataset and sample size.
Additionally, we conduct new ablation experiments to investigate the impact of different training datasets on performance, as presented in Table~\ref{Tab:1}. Our model's performance is equally good when trained on DAVIS. However, when trained on YouTube-VOS, we observe a significant degradation in performance mainly due to differences between the validation video dataset and the training data.

\subsubsection{Impact of Network Architecture}
To fulfill the input requirements of HCPN, we use a three-stream network. We investigate how the parallel encoder architecture of two consecutive frames contributes to performance through a quantitative study on the stream structure. As shown in Table~\ref{Tab:1}, the use of the three-stream encoder enhances performance (81.2\% $\to$ 85.4\% on Mean $\mathcal{F}$) compared to the two-stream encoder.

\subsubsection{Efficacy of Feature Fusion}
To obtain more abundant spatio-temporal features, two consecutive frames, each with its corresponding optical flow passing through PCM, require feature fusion. We propose CCM for cross-modal feature fusion, which leads to a 3.1\% improvement on Mean $\mathcal{J}$, as shown in Table~\ref{Tab:2}. Furthermore, compared to commonly used addition and concatenation operations, as presented in Table~\ref{Tab:1}, CCM provides an improvement of 2.6\% and 2.2\% on Mean $\mathcal{J}$. Additionally, the GAC module enhances the fusion features in the bridge network and improves Mean $\mathcal{J}$ by 1.1\%.
Moreover, we perform an ablation study using PWC~\cite{PWC} to investigate the impact of different optical flow estimation methods on the proposed model. As illustrated in Table~\ref{Tab:1}, while better optical flow estimation typically leads to improved segmentation accuracy, our model exhibits significant robustness against the quality of optical flow estimation.

\subsubsection{Benefit of MCR}
We have obtained compelling foreground-related features through propagation using PCM and CCM. However, the stationary background will shift as the camera moves. In our decoder network, we investigate the effect of the MCR module on the coarse masks. Table~\ref{Tab:2} reports the helpfulness of the MCR in optimizing motion contours and automatically identifying hard negative pixels.

\subsubsection{Qualitative Results and Attribute-Based Analysis}
It's great to see that the performance of the full HCPN model is compared to its different variants on sequence~\textit{drift-chicane} from DAVIS-16. The visualization results in Fig.~\ref{fig:6} will help to understand the effectiveness of the different components in capturing the foreground objects accurately. Moreover, the evaluation and summary of the HCPN components based on attribute analysis in Fig.~\ref{fig:7} will provide a better understanding of the model's strengths and weaknesses in handling different challenges in video object segmentation. This will also help in identifying the areas of improvement for future research.

\begin{table*}[t]
	\caption{The quantitative evaluation on the test set of YouTube-Objects over Mean $\mathcal{J} \uparrow$ (see \S{\ref{sec:4.3.1}} for details).}\label{Tab:5}
	\begin{center}
		\begin{tabular}{|r||ccccccccccccc|}
			\hline
			\multirow{2}*{Methods~~~} &MOT &LSMO &LVO &FSEG &PDB &SFL &AGS &COSNet &AGNN &MATNet &AMCNet &TMO &Ours \\
			~ &\cite{MOT} &\cite{LSMO} &\cite{LVO} &\cite{FSEG} &\cite{PDB} &\cite{SFL} &\cite{AGS} &\cite{COSNet} &\cite{AGNN} &\cite{MATNet} &\cite{AMCNet} &\cite{TMO} & \\
			\hline\hline
			Airplane(6) &77.2 &60.5 &\mkb{86.2} &81.7 &78.0 &65.6 &\mkr{87.7} &81.1 &81.1 &72.9 &78.9 &\mkg{85.7} &84.5 \\
			Bird(6) &42.2 &59.3 &\mkr{81.0} &63.8 &\mkg{80.0} &65.4 &76.7 &75.7 &75.9 &77.5 &\mkb{80.9} &\mkg{80.0} &79.6 \\
			Boat(15) &49.3 &62.1 &68.5 &\mkr{72.3} &58.9 &59.9 &\mkb{72.2} &\mkg{71.3} &70.7 &66.9 &67.4 &70.1 &67.3 \\
			Car(7) &68.6 &72.3 &69.3 &74.9 &76.5 &64.0 &78.6 &77.6 &78.1 &\mkg{79.0} &\mkb{82.0} &78.0 &\mkr{87.8} \\
			Cat(16) &46.3 &66.3 &58.8 &68.4 &63.0 &58.9 &69.2 &66.5 &67.9 &\mkb{73.7} &69.0 &\mkg{73.6} &\mkr{74.1} \\
			Cow(20) &64.2 &67.9 &68.5 &68.0 &64.1 &51.2 &64.6 &\mkg{69.8} &69.7 &67.4 &69.6 &\mkb{70.3} &\mkr{71.2} \\
			Dog(27) &66.1 &70.0 &61.7 &69.4 &70.1 &54.1 &73.3 &\mkb{76.8} &\mkr{77.4} &75.9 &75.8 &\mkb{76.8} &\mkg{76.5} \\
			Horse(14) &64.8 &65.4 &53.9 &60.4 &\mkr{67.6} &64.8 &64.4 &\mkb{67.4} &\mkg{67.3} &63.2 &63.0 &66.2 &66.2 \\
			Motorbike(10) &44.6 &55.5 &60.8 &62.7 &58.4 &52.6 &62.1 &\mkb{67.7} &\mkr{68.3} &62.6 &63.4 &58.6 &\mkg{65.8} \\
			Train(5) &42.3 &38.0 &\mkr{66.3} &\mkb{62.2} &35.3 &34.0 &48.2 &46.8 &47.8 &51.0 &57.8 &47.0 &\mkg{59.7} \\
			\hline
			Avg. &58.1 &64.3 &67.5 &68.4 &65.5 &57.1 &69.7 &70.5 &70.8 &69.0 &\mkg{71.1} &\mkb{71.5} &\mkr{73.3} \\
			\hline
		\end{tabular}
	\end{center}
\end{table*}

\subsection{Comparison with State-of-the-arts}\label{sec:4.3}

\subsubsection{Object-Level ZS-VOS}\label{sec:4.3.1}

\textbf{Validation set of DAVIS-16.} 
It is impressive to see that the proposed HCPN outperforms all previous top-performing methods on the DAVIS-16 validation set. Specifically, HCPN improves by 1.0\% on Mean $\mathcal{J}$\&$\mathcal{F}$ compared to AMCNet~\cite{AMCNet}, and 4.1\% and 4.7\% in terms of Mean $\mathcal{J}$ and Mean $\mathcal{F}$, respectively, compared to MATNet~\cite{MATNet}, which uses the same encoder-decoder structure and an additional dataset YouTube-VOS. Moreover, HCPN achieves better performance than 3DCSEG~\cite{3DCSEG}, which is pre-trained on IG-65M~\cite{IG-65M} + Kinetics~\cite{Kinetics} and fine-tuned on COCO~\cite{COCO}, YouTube-VOS~\cite{YouTube-VOS}, and DAVIS-16~\cite{DAVIS-16}, using fewer data. Table~\ref{Tab:4} shows detailed quantitative results for comparison.

\begin{table}[t]
	\footnotesize
	\caption{The quantitative evaluation on the test set of FBMS over Mean $\mathcal{J}$ (see \S{\ref{sec:4.3.1}} for details).}\label{Tab:6}
	\begin{center}
		\begin{tabular}{|c||ccccc|}
			\hline
			\multirow{2}*{Methods} &LSMO &PDB &LVO &IET &SFL \\
			~ &\cite{LSMO} &\cite{PDB} &\cite{LVO} &\cite{IET} &\cite{SFL} \\
			\hline
			Mean $\mathcal{J}$ &72.4 &72.3 &64.7 &71.9 &56.0 \\
			\hline\hline
			\multirow{2}*{Methods} &FSEG &FST &AGS &COSNet &MATNet\\
			~ &\cite{FSEG} &\cite{FST} &\cite{AGS} &\cite{COSNet} &\cite{MATNet} \\
			\hline
			Mean $\mathcal{J}$ &68.4 &55.5 &76.0 &75.6 &76.1 \\
			\hline\hline
			\multirow{2}*{Methods}  &3DCSEG &AMCNet & IMP & PMN &Ours \\
			~ &\cite{3DCSEG} &\cite{AMCNet} &\cite{IMP} &\cite{PMN} & \\
			\hline
			Mean $\mathcal{J}$ &76.2 &76.5 &\mkg{77.5} &\mkb{77.8} &\mkr{78.3} \\
			\hline
		\end{tabular}
	\end{center}
\end{table}

\textbf{Test set of YouTube-Objects.}
YouTube-Objects comprises ten categories that effectively evaluate the robustness of proposed VOS methods. Each category presents unique challenges, as evidenced by the average ``Mean $\mathcal{J}$" score across all categories (\textit{i.e.}, Avg. in Table~\ref{Tab:5}).
HCPN is a category-agnostic method and thus can be applied universally across different object categories. Table~\ref{Tab:5} provides detailed category-level results for YouTube-Objects, where our proposed method outperforms other methods in six of the ten categories. Notably, our HCPN achieves surprising results for both fast-moving and slow-moving objects, thanks to the synergistic cooperation of optical flow and continuous frames, allowing motion and appearance features to complement each other.
Regarding the average category metric, our method outperforms all other methods, with a 1.8\% improvement over TMO~\cite{TMO}.

\textbf{Test set of FBMS.}
Optical flow-based VOS methods often fail when target objects exhibit diverse motion patterns, which presents a significant challenge. To address this issue, we propose a combination of continuous frames and optical flow. Our method relies on appearance information between two consecutive frames to segment primary objects when optical flow estimation cannot express target motion information. We directly evaluate our proposed method on FBMS~\cite{FBMS} with video clips featuring multi-motion patterns (fast, slow, and stationary). As shown in Table~\ref{Tab:6}, our HCPN achieves the best results, with 78.3\% over Mean $\mathcal{J}$, outperforming the second-best method, PMN~\cite{PMN}, by 1.5\%. Quantitative results demonstrate that HCPN still delivers satisfactory segmentation performance when objects have complex motion patterns.

\begin{figure}[t]
	\begin{center}
		\includegraphics[width=0.95\linewidth]{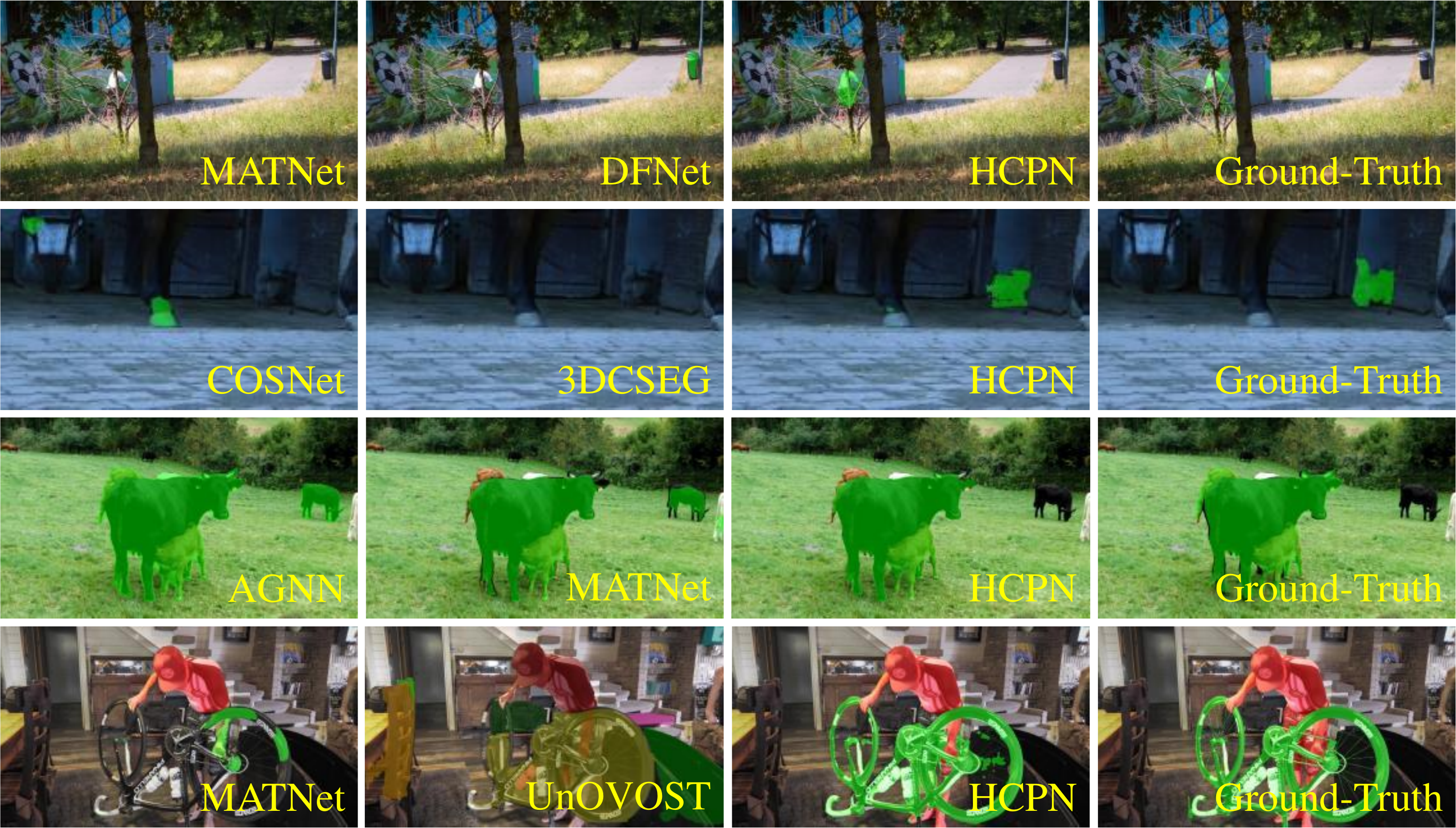}
	\end{center}
	\caption{Qualitative results on four example video sequences (see \S{\ref{sec:4.4.1}} for details). From top to bottom: \textit{bmx-trees} from DAVIS-16, \textit{dogs01} from FBMS, \textit{cow-0001} from YouTube-Objects, \textit{bike-packing} from DAVIS-17.}\label{fig:8}
\end{figure}

\begin{figure*}[t]
	\begin{center}
		\includegraphics[width=1.0\linewidth]{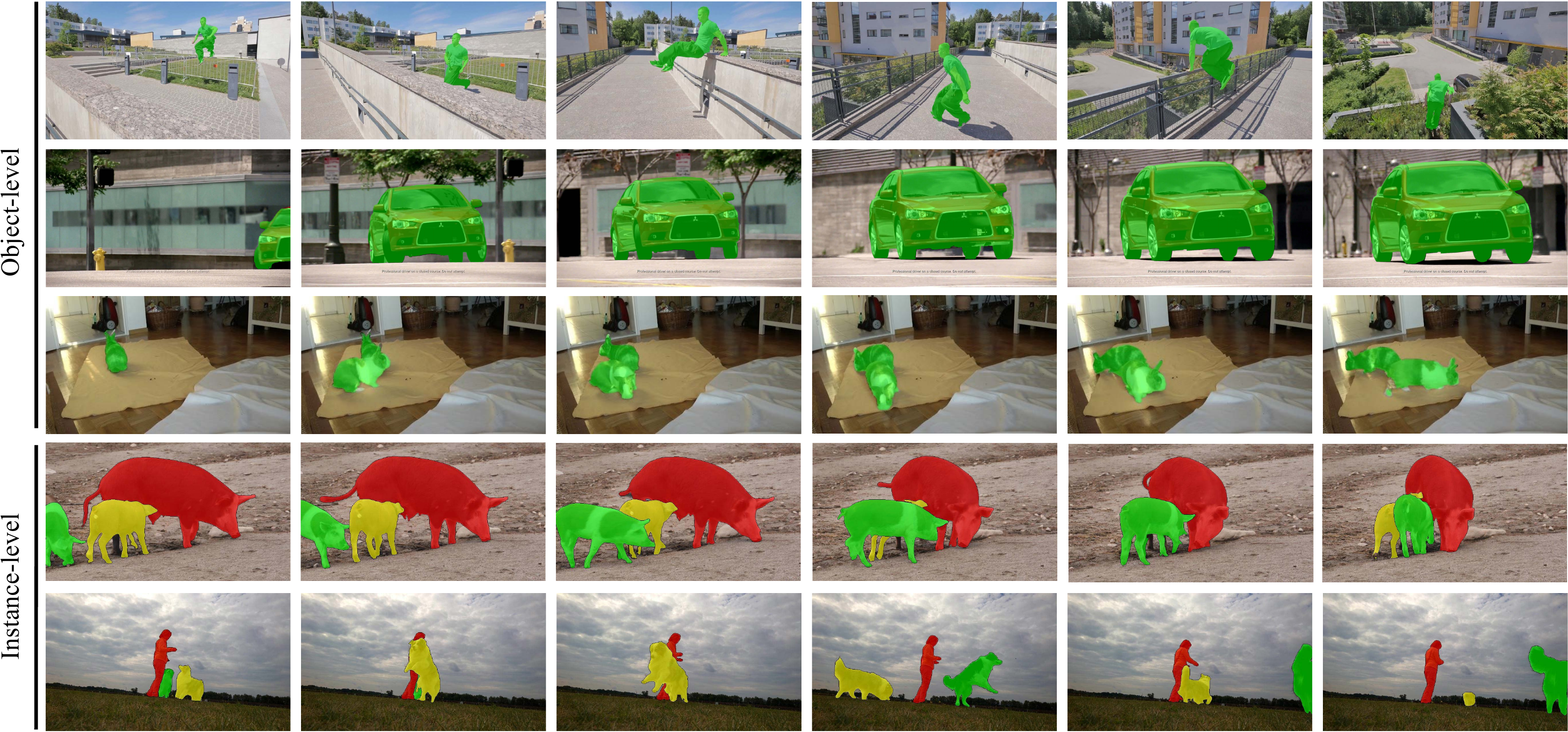}
	\end{center}
	\caption{Qualitative results on five example video sequences (see \S{\ref{sec:4.4.2}} for details). From top to bottom: \textit{parkour} from DAVIS-16, \textit{car-0004} from YouTube-Objects, \textit{rabbits-02} from FBMS, \textit{pigs} and \textit{dogs-jump} from DAVIS-17.}\label{fig:9}
\end{figure*}

\begin{table*}[t]
	\caption{The quantitative evaluation on the validation set of DAVIS-17 (see \S{\ref{sec:4.3.2}} for details). The results for other methods are borrowed from the public leaderboard.}\label{Tab:7}
	\begin{center}
		\begin{tabular}{|r||c||c||ccc||ccc||c|}
			\hline
			\multirow{2}*{Methods~~~~~~} &\multirow{2}*{Publications} &\multirow{2}*{Backbone} &\multicolumn{3}{c||}{$\mathcal{J}$} &\multicolumn{3}{c||}{$\mathcal{F}$} &$\mathcal{J}$\&$\mathcal{F}$ \\
			~ & ~ & ~ &Mean$\uparrow$ &Recall $\uparrow$ &Decay$\downarrow$ &Mean$\uparrow$ &Recall$\uparrow$ &Decay$\downarrow$ &Mean$\uparrow$ \\
			\hline\hline
			PDB~\cite{PDB}  &ECCV-2018 &ResNet-50 &53.2 &58.9 &4.9 &57.0 &60.2 &6.8 &55.1 \\
			RVOS~\cite{ventura2019rvos}  &CVPR-2019 &ResNet-101 &36.8 &40.2 &0.5 &45.7 &46.4 &1.7 &41.2 \\
			ALBA~\cite{gowda2020alba}  &BMVC-2020 &- &56.6 &63.4 &7.7 &60.2 &63.1 &7.9 &58.4 \\
			MATNet~\cite{MATNet_TIP} &TIP-2020 &ResNet-101 &56.7 &65.2 &\mkb{-3.6} &60.4 &68.2 &1.8 &58.6 \\
			STEm-Seg~\cite{athar2020stem}  &ECCV-2020 &ResNet-101 &61.5 &\mkg{70.4} &\mkr{-4} &67.8 &\mkg{75.5} &\mkb{1.2} &64.7 \\
			UnOVOST~\cite{UnOVOST}  & WACV-2020 &ResNet-101 &\mkb{66.4} &\mkb{76.4} &\mkg{-0.2} &\mkg{69.3} &\mkb{76.9} &\mkr{0.01} &\mkg{67.9} \\
			AGS~\cite{AGS_TPAMI}  &TPAMI-2021 &ResNet-101 &55.5 &61.6 &7.0 &59.5 &62.8 &9.0 &57.5 \\
			ProposeReduce~\cite{PRVIS} &ICCV-2021 &ResNet-101 &\mkg{65.0} &- &- &\mkb{71.6} &- &- &\mkb{68.3} \\
			D$^2$Conv3D~\cite{D2Conv3D} &WACV-2022 &ResNet-101 &60.8 &- &- &68.5 &- &- &64.6 \\
			\hline
			Ours &- &ResNet-101 &\mkr{68.7} &\mkr{77.7} &0.9 &\mkr{72.7} &\mkr{80.3} &\mkg{1.5} &\mkr{70.7} \\
			\hline
		\end{tabular}
	\end{center}
\end{table*}

\subsubsection{Instance-Level ZS-VOS}\label{sec:4.3.2}
In contrast to object-level ZS-VOS, instance-level ZS-VOS requires not only efficient identification of primary objects but also the differentiation of instance labels in the same region. To address this challenge, we adapt HCPN and validate it on the benchmark dataset DAVIS-17~\cite{DAVIS-17}.
Table~\ref{Tab:7} presents the quantitative performance results, which show that HCPN outperforms the nine state-of-the-art methods by a large margin. Specifically, our HCPN achieves a 2.4\% improvement in terms of Mean $\mathcal{J}$\&$\mathcal{F}$ compared to the second-best method, ProposeReduce~\cite{PRVIS}, with Mean $\mathcal{J}$ and Mean $\mathcal{F}$ outperforming by 3.7\% and 1.1\%.

For instance-level ZS-VOS, like existing proposal methods, our HCPN relies on accurate foreground prediction to achieve advanced results. To further investigate the advantages of HCPN in determining the main object regions, we conducted an ablation study in Table~\ref{Tab:8}, utilizing the same proposal approach (see \S{\ref{sec:4.1.2}} for details) in combination with previous object-level methods. HCPN outperforms other approaches in the instance-level ZS-VOS setting.

\begin{table}[t]
	\caption{Ablation study of various foreground proposal generation methods on DAVIS-17 validation set.}\label{Tab:8}
	\begin{center}
		\begin{tabular}{|c||c||cc||cc|}
			\hline
			Method &Variant &Mean $\mathcal{J}\uparrow$ &$\Delta \mathcal{J}$ &Mean $\mathcal{F}\uparrow$ &$\Delta \mathcal{F}$\\
			\hline\hline
			\multirow{3}*{Proposal} &MATNet &56.2 &-12.5 &60.7 &-12\\
					&3DCSEG &65.5 &-3.2 &69.9 &-2.8\\
					&Ours &68.7 &- &72.7 &-\\
			\hline
		\end{tabular}
	\end{center}
\end{table}

\subsection{Qualitative Results}\label{sec:4.4}

\subsubsection{Qualitative Comparisons of ZS-VOS Methods}\label{sec:4.4.1}

We select representative video sequences from each of the four datasets and introduce two high-performance comparison methods for each dataset. Fig.~\ref{fig:8} illustrates the qualitative results from top to bottom for DAVIS-16~\cite{DAVIS-16}, FBMS~\cite{FBMS}, YouTube-Objects~\cite{YouTube-Objects}, and DAVIS-17~\cite{DAVIS-17}. Our proposed HCPN achieves accurate segmentation results for fast- or slow-moving objects and object- or instance-level tasks.

\subsubsection{Visual Results of Challenging Video Sequences}\label{sec:4.4.2}

Fig.~\ref{fig:9} presents some qualitative results of the proposed HCPN on challenging videos from different datasets. The selected frames contain various challenges, such as fast-motion, occlusion, and out-of-view. The visual results show that our HCPN can achieve accurate and robust segmentation masks on both foreground objects and complex backgrounds. Furthermore, in the instance-level ZS-VOS setting, the proposed method can also produce high-quality instance masks. The qualitative results demonstrate the effectiveness of HCPN in addressing object-level and instance-level tasks.

\begin{figure}[t]
	\begin{center}
		\includegraphics[width=1\linewidth]{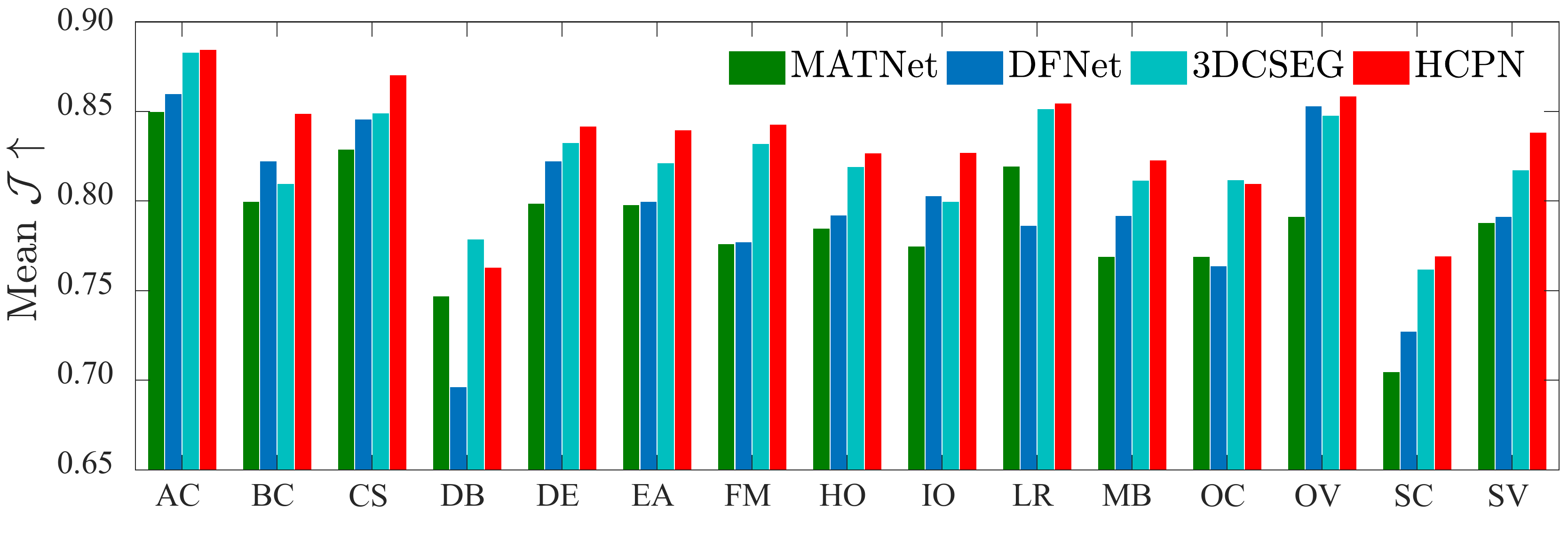}
	\end{center}
	\caption{Attribute-based study on DAVIS-16 (see \S{\ref{sec:4.5}} for details).}\label{fig:10}
\end{figure}

\subsection{Attribute-Based Study}\label{sec:4.5}
Table~\ref{Tab:4} and Fig.~\ref{fig:10} show the results of an attribute-based analysis on the DAVIS-16~\cite{DAVIS-16} validation set, comparing the proposed HCPN method with three other ZS-VOS methods, MATNet~\cite{MATNet}, DFNet~\cite{DFNet}, and 3DCSEG~\cite{3DCSEG}. HCPN outperforms the other methods in many attributes, including BC, CS, DE, EA, FM, HO, IO, MB, SC, and SV. These results suggest that HCPN is a robust approach that can effectively address various challenges in ZS-VOS.

\section{Conclusion}

This paper presents a novel hierarchical co-attention propagation network (HCPN) to automatically explore inter- and intra-frame representations of motion and appearance features. Experimental results on various challenging datasets show that HCPN outperforms existing state-of-the-art methods in object- and instance-level tasks. This is attributed to HCPN's ability to complement and enhance motion and appearance information, allowing for a good representation of spatio-temporal features throughout the video. Additionally, the multi-scale contour constraints between adjacent frames facilitate the determination of an object's relative motion to the background.

While HCPN improves the accuracy of primary object localization for ZS-VOS, some limitations of our study should be noted. The computational cost of our proposed approach is still too high to meet real-time requirements, and the cumulative errors introduced by optical flow estimation could potentially impair the performance of ZS-VOS. To address these limitations, future research could explore the design of lightweight ZS-VOS models and investigate alternatives to optical flow estimation. These directions could build upon our current findings and further advance the field.

Overall, HCPN is a general-purpose scheme for representing video sequence information that can be readily applied to other video understanding tasks, such as video classification, motion recognition, and video salient object segmentation.

{\small
	\bibliographystyle{IEEEtran}
	\bibliography{egbib}
}

\vfill

\end{document}